\documentclass[preprint, numafflabel, 12pt]{elsarticle}
\usepackage{graphicx}
\usepackage{float}

\usepackage[linesnumbered,ruled,vlined, algo2e]{algorithm2e}
\usepackage[utf8]{inputenc}
\usepackage{float}
\usepackage[section]{placeins}
\usepackage{graphicx}
\usepackage{subcaption}

\usepackage{amssymb}
\usepackage{amsmath}

\journal{Journal of Computers & Graphics}

\begin{document}

    \begin{frontmatter}

    \title{Synthetic data generation framework for quality control automation in rotogravure printing\tnoteref{t1}}
    
    \author[1,2]{Korota Arsène COULIBALY\corref{cor1}}
    \ead{korota.colibaly-etu@etu.univh2c.ma}
    \ead[Code]{https://github.com/Korotaa/STAGE-PLASTIMA}
    
    \author[1]{Mohamed HAMLICH}
    
    \author[2]{Khalid HMALI}
    
    \author[2]{Andrea TROMBIN}

    
    \cortext[cor1]{Corresponding author.}
    
    \affiliation[1]{organization={LCCPS Lab, ENSAM, Hassan II University of Casablanca},
               addressline={150 Bd du Nil},
                city={Casablanca},
                postcode={20670}, 
                state={},
                country={Morocco}}
    
    \affiliation[2]{organization={Plastima},
                addressline={Commune Chellalat, Route Secondaire 3002, Mohammedia},
                city={Casablanca},
                postcode={}, 
                state={},
               country={Morroco}}

    \begin{abstract}
    Quality control in printing, particularly in rotogravure printing, still depends on slow, costly, and subjective manual inspection. Automated surface defect detection is critical for maintaining high-quality standards in rotogravure printing. Deep learning models give prospects for automation. However, training robust deep learning models, such as YOLO or Vision Transformers, is heavily hindered by the extreme scarcity of real-world industrial defects images. To overcome this limitation, this paper introduces a novel synthetic data generation framework tailored for rotogravure printing quality control. The proposed pipeline automatically generates high-fidelity images of specific printing defects (creases, streaks, misregistration, etc.) and outputs corresponding bounding boxes and annotations. To validate the framework, a synthetic dataset of 7533 images was generated and used to train the state-of-the-art object-detection model RFDETR. Experimental results demonstrate that the model trained on our synthetic data achieves a Mean Average Precision (mAP) of 80.9\% on real industrial testing samples. This framework provides a zero-cost, rapid-deployment solution for automating defect inspection in printing lines without requiring massive manual data collection.

    \end{abstract}
    
    


    \begin{keyword}
    synthetic data \sep semantic segmentation \sep gravure printing defects \sep computer vision, \sep quality control \sep Industry 4.0 
    
    
    \end{keyword}
    
    \end{frontmatter}
    
    
    
    \section{Introduction}
    
    Maintaining good print quality is a fundamental pillar of the gravure printing industry. It directly impacts customer satisfaction and the company's reputation. Traditionally, quality control is based on visual inspection by expert operators on dedicated machines called viewing machines. This manual process has inherent limitations: it is slow, labor-intensive, and subject to operator subjectivity and tiredness. The advent of Industry 4.0, computer vision, and deep learning, particularly semantic segmentation models, offers an unprecedented opportunity to automate and objectify this inspection \cite{villalba-diezDeepLearningIndustrial2019}, \cite{wu2026weakly}.
    
    These technologies not only detect the presence of an anomaly in an image, but also locate and classify it with pixel-level accuracy, providing crucial information for the diagnosis of root causes \cite{chenTransUNetTransformersMake2021},\cite{chenRethinkingAtrousConvolution2017},\cite{hatamizadehUnetrTransformers3d2022}. A real-time defect detection system presented in \cite{shankarRealtimePrintdefectDetection2009}, analyzes print images captured by CCD cameras, scanned via a frame grabber, and compared to a gold master image with no defects. Defect detection is based on the identification of Regions of Interest (ROI) by contour detection algorithms based on pixel intensity differences. A correlation subtraction ensures image synchronization for accurate comparison. Defective pixels are grouped by correlation analysis and then classified according to their size and structure. However, the deployment of these computer vision systems is confronted with a major obstacle, well-known in the field of industrial vision: the difficulty in obtaining sufficient training data \cite{ahmadDeepLearningMethods2022}. The creation of a sufficiently large, diversified, and, especially, precisely labeled dataset at the pixel level in gravure printing is a very complex process, often due to the very small nature of defects (fisheyes). Collecting thousands of examples of real defects, which are by nature rare events, and annotating them manually and accurately represents an investment in time and resources that is often incompatible with the constraints of an industrial project (deadlines, budgets). This challenge is amplified by another reality of our production: the constant variability of printed patterns to meet customer orders. This dynamic makes anomaly detection approaches based on comparison to a reference pattern or AI models that learn all patterns extremely complex to implement, constantly changing the reference model. The need for a flexible system capable of adapting to this continuous flow of new designs is therefore essential. To respond to these challenges, this article presents a comprehensive framework for synthetic data generation, designed to enable the training of an automated, accurate, and flexible computer vision quality control system.
    
    The main contributions of this work are as follows:
    \begin{enumerate}
    
        \item A multi-class print defect simulation framework. We propose a methodology for modeling and generating a diverse range of physically realistic defects (streaks, creases, misregistration, etc.), enabling large-scale dataset creation for deep Convolutional Neural Networks (CNN) systems.
        \item An automatic labeling method for semantic segmentation. Our approach generates defective images and their perfect pixel-level segmentation masks, eliminating the need for costly and error-prone manual annotation.
    \end{enumerate}
    
    This paper is structured as follows: Section 2 (Materials and Methods) provides an overview of the related works and describes our innovative framework for generating synthetic data. The simulation results are presented in Section 3 (Results). Section 4 (Discussion) provides an in-depth analysis of these results, exploring their implications for the gravure printing industry, current constraints, and potential improvements, before concluding in Section 5.

    \section{Material and methods}
    \label{sec1}
    
    Our framework is applied in the field of rotogravure printing or roll-to-roll printing. This is an industrial printing technique that involves transferring gravure-printed ink onto a continuous substrate, such as paper, plastic, or metal film, which is rolled from one cylinder to another. It is widely used in areas such as packaging, decoration, publishing, and printed electronics, where precision and consistency are essential \cite{nohScalabilityRolltoRollGravurePrinted2010}.
    
        \subsection{Related Works}

        \cite{urgoMonitoringManufacturingSystems2024} presents a methodology that uses a factory's digital twin to generate synthetic data for training  CNNs, enabling the simulation of various scenarios and increasing the diversity of data available for online quality control. Similarly, in \cite{ngPrintinginspiredDigitalTwin2024} authors introduce MicroFactory, an autonomous digital twin that generates synthetic data to optimize the production of photovoltaic cells printed with a roll-to-roll printer. Their approach combines high-throughput device manufacturing with the use of machine learning models to simulate a wide range of production configurations, facilitating the identification of optimal parameters and illustrating the potential of synthetic data for the optimization of complex processes.
        In \cite{valentePrintDefectMapping2020a}, the authors developed a pioneering approach to pixel-level defect mapping. Their framework, based on semantic segmentation with DeepLab-v3+, was characterized by an innovative method of synthetic data generation. Faced with the scarcity of real annotated datasets, they synthesized dark streaks and banding defects by integrating advanced techniques for optical and color reconstruction. More specifically, the simulation of defects included the insertion of textured dark streaks whose colorimetric intensity varied through the application of Perlin noise, as well as the superimposition of rectangular color bands obtained by manipulating the CMYK (Cyan, Magenta, Yellow, Black) channels and modeling pixel intensity variations via a bimodal Gaussian distribution. To accurately reproduce the effect of printing and scanning, they calibrated the RGB (Red, Green, Blue) tone curves of actual devices and established a color transformation model using linear least squares regression.

        This study extends and adapts these principles of synthetic data generation specifically to the inspection of defects in gravure printing on roll-to-roll printers. The modified image and its corresponding segmentation mask are generated simultaneously. We focus on the pixel-level identification and mapping of a specific range of defects critical to this type of production, including misregistrations, fisheyes, streaks, and creases.  This approach allows us to quickly build a large, varied dataset (we control all parameters) that is accurately annotated, which is essential for training a segmentation model. The goal is to leverage the benefits of semantic segmentation and the flexibility of synthetic data generation to develop high-precision, fully automated inspection systems, which are essential in a challenging industrial environment.

        \subsection{Synthetic data generation Framework}
       
        This framework is a direct response to the constraints of the project. The difficulty lies in acquiring and labeling a sufficient volume of real data within the deadline (five months). By modeling defects, we turn this constraint into an advantage. We have complete control over the variety and quantity of data generated. We obtain perfect labels because the mask for each defect is created at the same time as the defect itself.
    
        To model gravure printing defects(misregistration, creases, streaks, and fisheyes) we drew inspiration from the work in 
        \cite{chenHybridModelingCompensation2025},\cite{kangMathematicalModelingSimulations2014},\cite{leeDatadrivenFaultDetection2024},\cite{zhangModelBasedPrecise2024},\cite{chenNonlinearWebTension2022} on cylinder dynamics and registration control in roll-to-roll systems and adapted it to our case study. Based on these principles, we have developed appropriate simulation models, combining the mechanical parameters of the cylinders and ink deposition, in order to generate realistic synthetic data for training computer vision systems and evaluating quality control strategies.
    
        \subsubsection{Framework architecture}
    
\begin{figure*}[t]
                \centering
                \includegraphics[width=\textwidth]{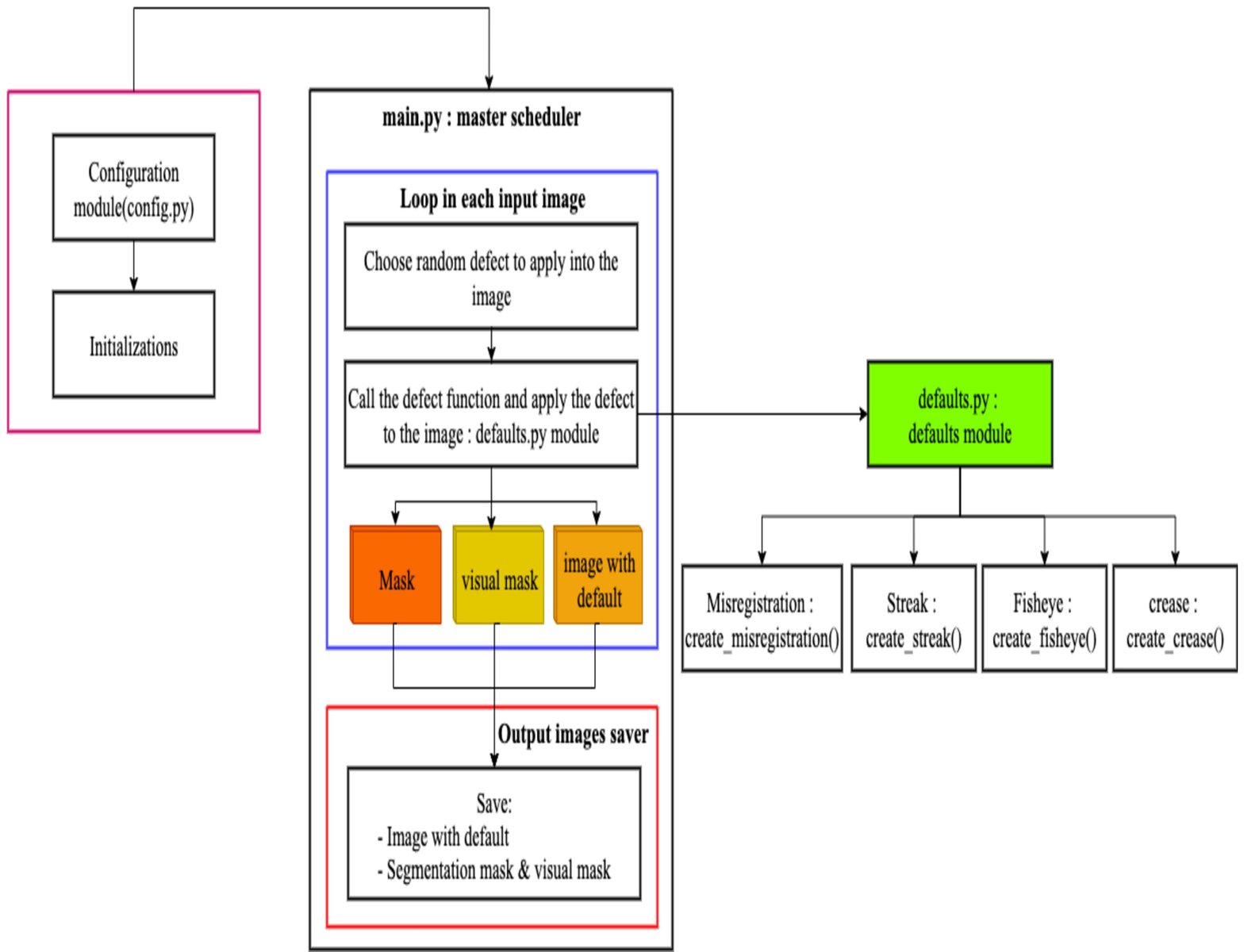}
                \caption{Framework architecture}\label{fig:Framework_architecture}
            \end{figure*}

        Our Framework shown in Figure \ref{fig:Framework_architecture}, operates in three steps:
        \begin{itemize}
            \item Setup: We configure the storage folders, the types of defects to simulate (number of defects per image, etc), and initialize the other necessary parameters.
            \item Defect generation: for each image, a random defect is applied. The framework then generates a segmentation mask (precise location of the defect), the image with the defect, and a visual mask (to facilitate visualization).
            \item Store: We constructed the synthetic dataset by storing all generated components, namely the defective images, segmentation masks, and visual masks.
        \end{itemize}

        \subsubsection{Framework algorithm} 
    
        The process, formalized in Algorithm 1, iterates over a set of healthy images to apply modeled defects to them. For each image, a random number of defects are applied sequentially. At each step, a simulation function (crease, shift, etc) is chosen at random with random parameters (size, position) to ensure diversity (lines 10-12). 
         
            \begin{algorithm2e}[h]
            \SetAlgoLined
            \DontPrintSemicolon
            
            \KwData{$D_{clean}$: Set of file paths to the input images}
            \KwData{$F_{defects}$: Set of defect simulation functions $\{f_1, f_2, ..., f_k\}$}
            \KwData{$N_{max}$: Maximum number of defects to apply per image}
            
            \KwResult{$D_{defective}$: Set of synthetic defective images}
            \KwResult{$M_{segmentation}$: Set of corresponding segmentation masks}
            
            \SetKwFunction{FGenDataset}{GenerateSyntheticDataset}
            \SetKwFunction{FLoadImg}{LoadImage}
            \SetKwFunction{FGetDim}{GetDimensions}
            \SetKwFunction{FCopy}{Copy}
            \SetKwFunction{FCreateZero}{CreateZeroMatrix}
            \SetKwFunction{FRandInt}{RandomInteger}
            \SetKwFunction{FChooseRandomly}{ChooseRandomly}
            \SetKwFunction{FGenRandParams}{GenerateRandomParameters}
            \SetKwFunction{FAdd}{Add}
            \SetKwProg{Fn}{Function}{}{end}
            \Fn{\FGenDataset{$D_{clean}, F_{defects}, N_{max}$}}{
                $D_{defective} \leftarrow \emptyset$\;
                $M_{segmentation} \leftarrow \emptyset$\;
                
                \ForEach{$image\_path$ \textbf{in} $D_{clean}$}{
                    $I_{original} \leftarrow \FLoadImg(image\_path)$\;
                    $height, width \leftarrow \FGetDim(I_{original})$\;
                    $I_{final} \leftarrow \FCopy(I_{original})$\;
                    $M_{final} \leftarrow \FCreateZero(height, width)$\;
                    
                    $n_{defects} \leftarrow \FRandInt(1, N_{max})$\;
                    
                    \For{$i \leftarrow 1$ \textbf{to} $n_{defects}$}{
                        $f_{selected} \leftarrow \FChooseRandomly(F_{defects})$\;
                        $parameters \leftarrow \FGenRandParams(f_{selected}, width, height)$\;
                        
                        \tcp{Apply the defect and retrieve the modified image and mask}
                        $I_{temp}, M_{temp} \leftarrow f_{selected}(I_{final}, parameters)$\;
                        
                        \tcp{Update the final image and the global mask}
                        $I_{final} \leftarrow I_{temp}$\;
                        $M_{final} \leftarrow M_{final} \lor M_{temp}$ \tcp*{Logical OR operation to merge masks}
                    }
                    
                    \FAdd{$D_{defective}, I_{final}$}\;
                    \FAdd{$M_{segmentation}, M_{final}$}\;
                }
                
                \KwRet{$D_{defective}, M_{segmentation}$}\;
            }
            \caption{Synthetic print defect dataset generation}
            \label{alg:data_generation_en}
            \end{algorithm2e}

        \subsubsection{fisheye defect} 
    
        These are small circular or oval spots resulting from incomplete ink deposit, usually caused by foreign contamination introduced during the calendering process. The white spot defect, or fisheye, is simulated by modeling the effect of a single chemical contamination on the printing surface. The algorithm modifies the original image $I_0$ by applying a crater effect to $N$ positions, thus generating a final defective image $I_f$. The effect of a single point of contamination centered at $(c_x, c_y)$ is described by the following equation, which alters the value of a pixel $I_0(x,y)$ according to its distance from the center of the defect:
        
        \begin{equation}
        I_f(x, y) = 
        \begin{cases} 
        C_{\text{ground}} & \text{if } d(x,y) \leq r_c \\
        I_0(x, y) \cdot \beta & \text{if } r_c < d(x,y) \leq r \\
        I_0(x, y) & \text{else}
        \end{cases}
        \label{eq:fisheye}
        \end{equation}
        
       Each term in equation \ref{eq:fisheye} represents a part of the crater effect:
        \begin{itemize}
            \item $I_f(x, y)$ is the pixel value of the modified image at the  $(x, y)$ position.
            \item $I_0(x, y)$ is the pixel value of the original image.
            \item $d(x,y)$ is the euclidean distance of the pixel
            $(x,y)$ from the center of the defect $(c_x, c_y)$, defined by $d(x,y) = \sqrt{(x-c_x)^2 + (y-c_y)^2}$.
            \item $r$ is the total radius of the defect.
            \item $r_c$ is the radius of the crater's bright center, typically $r_c \approx 0.7 \cdot r$.
            \item $C_{\text{ground}}$ is the average color of the image background, simulating a complete lack of ink.
            \item $\beta$ is the darkness factor for the ink bead, with $\beta < 1$ (e.g., 0.85).
        \end{itemize}
        
        For each contamination point, the algorithm defines a circular area. If a pixel is in the inside circle (the crater center, $d \leq r_c$), its color is replaced by the background color. If it's in the outside circle (the rim, $r_c < d \leq r$), its color is darkened. Pixels outside the circle are not affected by this defect. The implementation of this model is detailed in Algorithm~\ref{alg:contamination}.
        
        \begin{algorithm2e}[H]
        \SetAlgoLined
        \DontPrintSemicolon 
        
        \KwData{$I_0$: The original RGB image}
        \KwData{$N_{\mathit{spots}}$: The number of defects to generate}
        \KwData{$R_{\mathit{range}}$: The range $[r_{\min}, r_{\max}]$ for the spot radius}
        \KwData{$C_{\mathit{value}}$: The integer class value for the segmentation mask}
        
        \KwResult{$I_f$: The final image with fisheye defects}
        \KwResult{$M_{\mathit{seg}}$: The corresponding segmentation mask}
        
        \SetKwFunction{FGenFisheye}{GenerateFisheyeDefect}
        \SetKwFunction{FConvertFloat}{ConvertToFloat}
        \SetKwFunction{FGetDim}{GetDimensions}
        \SetKwFunction{FCreateZero}{CreateZeroMatrix}
        \SetKwFunction{FCalcBG}{CalculateBackgroundColor}
        \SetKwFunction{FRandInt}{RandomInteger}
        \SetKwFunction{FCreateCirc}{CreateCircleMask}
        \SetKwFunction{FCreateRing}{CreateRingMask}
        \SetKwFunction{FSetValueWhere}{SetValueWhere}
        \SetKwFunction{FMultValueWhere}{MultiplyValueWhere}
        \SetKwFunction{FSetValueInMask}{SetValueInMask}
        \SetKwFunction{FClipConvert}{ClipAndConvertToUInt8}
        
        \SetKwProg{Fn}{Function}{}{end}
        \Fn{\FGenFisheye{$I_0, N_{\mathit{spots}}, R_{\mathit{range}}, C_{\mathit{value}}$}}{
            $I_f \leftarrow \FConvertFloat(I_0)$\;
            $h, w \leftarrow \FGetDim(I_0)$\;
            $M_{\mathit{seg}} \leftarrow \FCreateZero(h, w)$\;
            $C_{\mathit{fond}} \leftarrow \FCalcBG(I_0)$\;
            
            \For{$i \leftarrow 1$ \textbf{to} $N_{\mathit{spots}}$}{
                $c_x \leftarrow \FRandInt(0, w - 1)$\;
                $c_y \leftarrow \FRandInt(0, h - 1)$\;
                $r \leftarrow \FRandInt(R_{\mathit{range},\min}, R_{\mathit{range},\max})$\;
                $r_c \leftarrow \lfloor r \times 0.7 \rfloor$\;
                $\beta \leftarrow 0.85$\;
                
                \tcp{Create masks for the defect areas}
                $M_{\mathit{center}} \leftarrow \FCreateCirc((c_x, c_y), r_c)$\;
                $M_{\mathit{ring}} \leftarrow \FCreateRing((c_x, c_y), r_c, r)$\;
                
                \tcp{Apply visual effects based on the equation}
                \FSetValueWhere{$I_f, M_{\mathit{center}}, C_{\mathit{fond}}$}\;
                \FMultValueWhere{$I_f, M_{\mathit{ring}}, \beta$}\;
                
                \tcp{Update the final segmentation mask}
                $M_{\mathit{spot}} \leftarrow M_{\mathit{center}} \lor M_{\mathit{ring}}$\;
                \FSetValueInMask{$M_{\mathit{seg}}, M_{\mathit{spot}}, C_{\mathit{value}}$}\;
            }
            
            $I_f \leftarrow \FClipConvert(I_f, 0, 255)$\;
            \KwRet{$I_f, M_{\mathit{seg}}$}\;
        }
        \caption{Fisheye Algorithm}
        \label{alg:contamination}
        \end{algorithm2e}

    
        \subsubsection{Streak defect}
        
        These are fine lines or undesirable streaks resulting from non-uniform ink distribution or mechanical failure of the cylinder or squeegee. The streak defect is modelled as a localized photometric alteration in a vertical band. To avoid the artificial appearance of a simple line of uniform color, our simulation combines three components to define the intensity of the darkness at each point.
        The final defective image $I_f$ is obtained by multiplying the pixel values of the original image $I_0$ by an attenuation factor $S(x,y)$ within the streak area.
        
        \begin{equation}
        I_f(x, y) = 
        \begin{cases} 
        I_0(x, y) \cdot (1 - S(x,y)) & \text{si } (x,y) \in \text{Area}_{\text{streak}} \\
        I_0(x, y) & \text{else}
        \end{cases}
        \label{eq:streak}
        \end{equation}
        
        The intensity factor of the stripe $S(x,y)$ is the product of three distinct functions:
        \begin{equation}
        S(x,y) = A \cdot P(x) \cdot \Pi(y)
        \label{eq:streak_intensity}
        \end{equation}
        
        Each term in equation \ref{eq:streak_intensity} models a physical characteristic of the streak:
        \begin{itemize}
            \item $A$ is the base intensity of the streak, a global parameter.
            \item $P(x)$ is the cross-profile function, which smooths the edges of the stripe. It is represented by a parabolic function that reaches its maximum at the center of the streak ($x = x_c$) and becomes zero at its edges. For a streak of width $W$ centered at $x_c$, it is defined by:
            \begin{equation*}
                P(x) = 1 - \left( \frac{2(x - (x_c - W/2))}{W} - 1 \right)^2
                \label{eq:cross_profile}
            \end{equation*}
            \item $\Pi(y)$ is the longitudinal texture function, which simulates the variation in ink density along the streak. It is generated by 1D Perlin noise, normalized between 0 and 1. This function gives the streak its organic and non-uniform appearance \cite{s24227358}.
        \end{itemize}
        This composite approach allows synthetic streaks to be generated with a high degree of realism, as detailed in Algorithm~\ref{alg:streak_defect}.
        
           \begin{algorithm2e}[H] 
            \SetAlgoLined
            \DontPrintSemicolon
            
            \KwData{$I_0$: The original RGB image}
            \KwData{$x_c$: The center x-coordinate of the streak}
            \KwData{$W$: The width of the streak}
            \KwData{$A$: The base intensity of the streak}
            \KwData{$C_{\mathit{value}}$: The integer class value for the segmentation mask}
            
            \KwResult{$I_f$: The final image with the streak defect}
            \KwResult{$M_{\mathit{seg}}$: The corresponding segmentation mask}
            
            \SetKwFunction{FGenStreak}{GenerateStreakDefect}
            \SetKwFunction{FConvertFloat}{ConvertToFloat}
            \SetKwFunction{FGetDim}{GetDimensions}
            \SetKwFunction{FCreateZero}{CreateZeroMatrix}
            \SetKwFunction{FGenPerlinOneD}{GeneratePerlinNoise1D}
            \SetKwFunction{FClipConvert}{ClipAndConvertToUInt8}
            
            \SetKwProg{Fn}{Function}{}{end}
            \Fn{\FGenStreak{$I_0, x_c, W, A, C_{\mathit{value}}$}}{
                $I_f \leftarrow \FConvertFloat(I_0)$\;
                $h, w \leftarrow \FGetDim(I_0)$\;
                $M_{\mathit{seg}} \leftarrow \FCreateZero(h, w)$\;
                
                \tcp{Step 1: Generate the longitudinal texture}
                $\Pi \leftarrow \FGenPerlinOneD(h)$ \tcp*{Returns a vector of size h}
                
                \tcp{Step 2: Apply the effect column by column}
                $x_{\mathit{start}} \leftarrow x_c - \lfloor W/2 \rfloor$\;
                $x_{\mathit{end}} \leftarrow x_c + \lceil W/2 \rceil - 1$\;
                
                \For{$x \leftarrow x_{\mathit{start}}$ \textbf{to} $x_{\mathit{end}}$}{
                    \If{$0 \leq x < w$}{
                        \tcp{Calculate the transversal profile}
                        $P_x \leftarrow 1 - \left( \frac{2(x - x_{\mathit{start}})}{W} - 1 \right)^2$\;
                        
                        \tcp{Calculate the final intensity vector for the column}
                        $\vec{S}_x \leftarrow A \cdot P_x \cdot \Pi$\;
                        
                        \tcp{Apply the darkening effect to the image column}
                        $I_f[:, x] \leftarrow I_f[:, x] \times (1 - \vec{S}_x)$\;
                        
                        \tcp{Update the segmentation mask}
                        $M_{\mathit{seg}}[:, x] \leftarrow C_{\mathit{value}}$\;
                    }
                }
                
                $I_f \leftarrow \FClipConvert(I_f, 0, 255)$\;
                \KwRet{$I_f, M_{\mathit{seg}}$}\;
            }
            \caption{Streak Algorithm}
            \label{alg:streak_defect}
        \end{algorithm2e}

        \subsubsection{Misregistration defect}

        Misregistration is caused by misalignment of the printing cylinders, resulting in incorrect color superposition. To simulate the misregistration defect in a physically coherent form, we developed a composite mathematical model. This model is based on the prediction results of a contour detection CNN \cite{bertasiusDeepEdgeMultiscaleBifurcated2015},\cite{weishenDeepContourDeepConvolutional2015} that we trained to detect the contours of any pattern with very high accuracy. This mathematical model defines the final defective image, denoted $I_f$(shown in equation  \ref{eq:misregistration_synthesized}), as a function of the original image, $I_0$. The approach consists of transparently superimposing one or more simulated ink streaks onto the base image, as described by the following synthesized equation:
        \begin{multline}
            I_f(x, y) = (1-\alpha)I_0(x,y) + 
            \alpha \sum_{i=1}^{N} C_{c_i} \cdot ( D_k \circ T_{\vec{v_i}} ) ( M_{\text{edges}} \wedge M_{\text{color}}(c_i) ) (x,y) 
            \label{eq:misregistration_synthesized}
        \end{multline}
        
        Each equation term in \ref{eq:misregistration_synthesized} corresponds to a specific step in our simulation algorithm, which mimics the actual printing process. We break down the equation as follows:
        
        \begin{itemize}
            \item\textbf{$I_0(x,y)$ and $\alpha$:} The term $I_0(x,y)$ represents the substrate (our PVC film) already printed correctly by the well-aligned cylinders. The term $\alpha \cdot (\dots)$ simulates the offset ink deposit, where $\alpha$ is the transparency factor of this superimposed defect layer.
        
            \item\textbf{$M_{\text{edges}} \wedge M_{\text{color}}(c_i)$:} This term represents the information scribed on a specific cylinder. To obtain it, we proceed in two steps. First, $M_{\text{edges}}$ identifies all the edges of the patterns in the final image. Second, $M_{\text{color}}(c_i)$ isolates the color area specific to a cylinder. The logical intersection ($\wedge$) of these two masks retains only the edges of the shapes that must be printed with the color of cylinder $c_i$.
        
            \item\textbf{$(D_k \circ T_{\vec{v_i}})$:} This operator simulates the mechanical defect and its consequences. $T_{\vec{v_i}}$ (Translation) simulates the mechanical shift itself, while $D_k$ (Dilation) simulates the slight spreading of the ink.
        
            \item\textbf{$\sum_{i=1}^{N} C_{c_i} \cdot (\dots)$:} This term simulates the multi-cylinder \cite{nohScalabilityRolltoRollGravurePrinted2010a} printing process. $C_{c_i}$ is the ink color, and $\sum$ (Sum) calculates and adds the fringes for each offset cylinder to obtain the complete defect layer.
        \end{itemize}
        
        The mathematical model breaks down the defect into its physical causes. It isolates the information from a cylinder, simulates its incorrect mechanical alignment, and applies the corresponding ink to the final image (see Algorithm \ref{alg:misregistration_summary} for the implementation).
        
            \begin{algorithm2e}[H] 
            \SetAlgoLined
            \DontPrintSemicolon
            
            \KwData{$I_0$: The original RGB image}
            \KwData{$S$: A list of shift tasks, where each task is $(c_i, \vec{v_i})$}
            \KwData{$M_{edge}$: The pre-trained edge detection model}
            
            \KwResult{$I_f$: The final image with the defect}
            \KwResult{$M_{seg}$: The corresponding segmentation mask}
            
            \SetKwFunction{FGenMisregistration}{GenerateMisregistrationDefect}
            \SetKwFunction{FGetDim}{GetDimensions}
            \SetKwFunction{FCopy}{Copy}
            \SetKwFunction{FCreateZero}{CreateZeroMatrix}
            \SetKwFunction{FExtractContours}{ExtractCleanContours}
            \SetKwFunction{FCreateFringe}{CreateShiftedFringe}
            \SetKwFunction{FAlphaBlend}{AlphaBlend}
            
            \SetKwProg{Fn}{Function}{}{end}
            \Fn{\FGenMisregistration{$I_0, S, M_{edge}$}}{
                $h, w \leftarrow \FGetDim(I_0)$ \tcp*{Get dimensions first}
                $I_f \leftarrow \FCopy(I_0)$\;
                $M_{seg} \leftarrow \FCreateZero(h, w)$\;
                $L_{combined} \leftarrow \FCreateZero(h, w, 3, \text{float})$\;
                
                \tcp{Step 1: Extract clean main contours from the entire image}
                $M_{contours} \leftarrow \FExtractContours(I_0, M_{edge})$\;
                
                \tcp{Step 2: Generate and accumulate fringes for each shift task}
                \For{each $(c_i, \vec{v_i})$ \textbf{in} $S$}{
                    \tcp{Isolate, shift, and paint the fringe for the current cylinder}
                    $L_{fringe}, M_{fringe} \leftarrow \FCreateFringe(I_0, M_{contours}, c_i, \vec{v_i})$\;
                    
                    $L_{combined} \leftarrow L_{combined} + L_{fringe}$\;
                    $M_{seg} \leftarrow M_{seg} \lor M_{fringe}$ \tcp*{Logical OR to merge masks}
                }
                
                \tcp{Step 3: Superimpose the combined fringes onto the original image}
                $I_f \leftarrow \FAlphaBlend(I_f, L_{\mathit{combined}}, \alpha)$\;
                
                \KwRet{$I_f, M_{\mathit{seg}}$}\;
            }
            \caption{Misregistration Algorithm}
            \label{alg:misregistration_summary}
        \end{algorithm2e}

        \subsubsection{Crease defect} 
        
        The crease defect is an anomaly that affects both the geometry of the pattern and its photometric appearance. Our simulation models both of these effects. The final defective image, $I_f$(equation \ref{eq:crease_main}), is obtained by applying a layer of visual effects $E$ to a geometrically distorted version of the original image,$I_d$.
        
        \begin{equation}
        I_f(x, y) = \text{clip}_{0}^{255} \left( I_d(x, y) + E(x, y) \right)
        \label{eq:crease_main}
        \end{equation}
        The defect image $I_d$, shown in equation \ref{eq:defaut_deplacement}, is obtained by applying a non-linear remapping function $R$ to the original image $I_0$, which simulates the pinching of the substrate towards the crease line.
        
        \begin{equation}
        I_d(x,y) = I_0\left[(x,y) + A_{\max} 
        \exp\left(-\frac{d(x,y)^2}{2\sigma^2}\right) 
        \hat{v}(x,y)\right],
        \label{eq:defaut_deplacement}
        \end{equation}

        Where:
        \begin{itemize}
            \item $A_{max}$ is the maximum amplitude of pixel displacement.
            \item $d(x, y)$ is the Euclidean distance from pixel $(x, y)$ to the fold line segment.
            \item $\exp\left(-\frac{d(x,y)^2}{2\sigma^2}\right)$ is a Gaussian function that ensures that the deformation force is maximum on the fold line and decreases with distance.
            \item $\hat{v}(x, y)$ is the unit vector pointing from pixel $(x, y)$ to the nearest point on the fold line, defining the direction of the pinch.
        \end{itemize}
        
        The effects layer $E(x, y)$ simulates the relief of the crease by adding shadows and reflections along a curved trajectory $P(t)$ for greater realism. For each point $p_i$ on this trajectory, brightness modifications are added to the $E$ layer. The intensity of these effects, $F(p_i)$, is modulated by the saturation of the underlying color and by Perlin noise for a more natural rendering as shown in.
        
        \begin {equation}
        F(p_i) = V \cdot \Pi(t_i) \cdot S(p_i)
        \end{equation}
        
        Where $V$ is a base intensity, $\Pi(t_i)$ is the Perlin noise value, and $S(p_i)$ is the color saturation at point $p_i$. Three effects are applied: a drop shadow, a specular reflection, and a center line simulating ink wear. The implementation of this model is detailed in Algorithm~\ref{alg:crease_defect}.
        
        \begin{algorithm2e}[H] 
                \SetAlgoLined
                \DontPrintSemicolon
                
                \KwData{$I_0$: The original RGB image}
                \KwData{$P_1, P_2$: The start and end points of the crease line}
                \KwData{$A_{max}$: The maximum pixel displacement for the distortion}
                \KwData{$\beta$: The brightness factor for the centerline}
                \KwData{$C_{value}$: The integer class value for the segmentation mask}
                
                \KwResult{$I_f, M_{seg}$: The final and segmentation mask images }
                
                \SetKwFunction{FGenCrease}{Crease defect} 
                \SetKwFunction{FGetDim}{GetDimensions}
                \SetKwFunction{FCreateIdentityRemap}{CreateIdentityRemap}
                \SetKwFunction{FCalcDistVec}{CalculateDistanceAndVectorToSegment}
                \SetKwFunction{FRemap}{Remap}
                \SetKwFunction{FCreateZero}{CreateZeroMatrix}
                \SetKwFunction{FGetSaturationMap}{GetSaturationMap}
                \SetKwFunction{FLength}{Length}
                \SetKwFunction{FGenPerlinOneD}{GeneratePerlinNoise1D}
                \SetKwFunction{FCalcWobblePath}{CalculateWobblePath}
                \SetKwFunction{FCalcEffectStrength}{CalculateEffectStrength}
                \SetKwFunction{FAddShadow}{AddShadowEffect}
                \SetKwFunction{FAddHighlight}{AddHighlightEffect}
                \SetKwFunction{FAddCenterline}{AddCenterlineEffect}
                \SetKwFunction{FClipConvert}{ClipAndConvertToUInt8}
                \SetKwFunction{FCreateMaskFromDistortion}{CreateMaskFromDistortion}

                \SetKwProg{Fn}{Function}{}{end}
                \Fn{\FGenCrease{$I_0, P_1, P_2, A_{max}, \beta, C_{value}$}}{
                    \tcp{Step 1: Geometric Distortion}
                    $h, w \leftarrow \FGetDim(I_0)$\;
                    $R_x, R_y \leftarrow \FCreateIdentityRemap(h, w)$\;
                    \For{each pixel $(x,y)$ in $I_0$}{
                        $d, \hat{v} \leftarrow \FCalcDistVec((x,y), P_1, P_2)$\;
                        $\vec{\Delta} \leftarrow A_{max} \cdot \exp(-d^2 / (2\sigma^2)) \cdot \hat{v}$\;
                        $R_x[y,x] \leftarrow x + \vec{\Delta}_x$\;
                        $R_y[y,x] \leftarrow y + \vec{\Delta}_y$\;
                    }
                    $I_d \leftarrow \FRemap(I_0, R_x, R_y)$\;
                    
                    \tcp{Step 2: Photometric Effects Layer}
                    $E \leftarrow \FCreateZero(h, w, 3, \text{float})$\;
                    $S_{map} \leftarrow \FGetSaturationMap(I_0)$\;
                    
                    $L_{path} \leftarrow \FLength(P_1, P_2)$ \tcp*{Decompose the operation}
                    $\Pi_{noise} \leftarrow \FGenPerlinOneD(L_{path})$\;
                    
                    $P_{path} \leftarrow \FCalcWobblePath(P_1, P_2, A_{max})$\;
                    
                    \For{each point $p_i$ at step $t_i$ along $P_{path}$}{
                        $F \leftarrow \FCalcEffectStrength(S_{map}[p_i], \Pi_{noise}[t_i])$\;
                        \FAddShadow{$E, p_i, F$}\;
                        \FAddHighlight{$E, p_i, F$}\;
                        \FAddCenterline{$E, p_i, F, \beta$}\;
                    }
                   \tcp{Step 3: Final Composition and Mask Generation}
                    $I_f \leftarrow I_d + E$\;
                    $I_f \leftarrow \FClipConvert(I_f, 0, 255)$\;
                    $M_{\mathit{seg}} \leftarrow \FCreateMaskFromDistortion(offset\_scale, C_{\mathit{value}})$\;
                    
                    \KwRet{$I_f, M_{\mathit{seg}}$}\;
                    
                  }
               
                \caption{Crease Algorithm}
                \label{alg:crease_defect}
                \end{algorithm2e}

    \section{Results} 
    
        \subsection{Experimental setup}
                           
\begin{figure}[H]   
            \centering
            \includegraphics[width=0.8\linewidth]{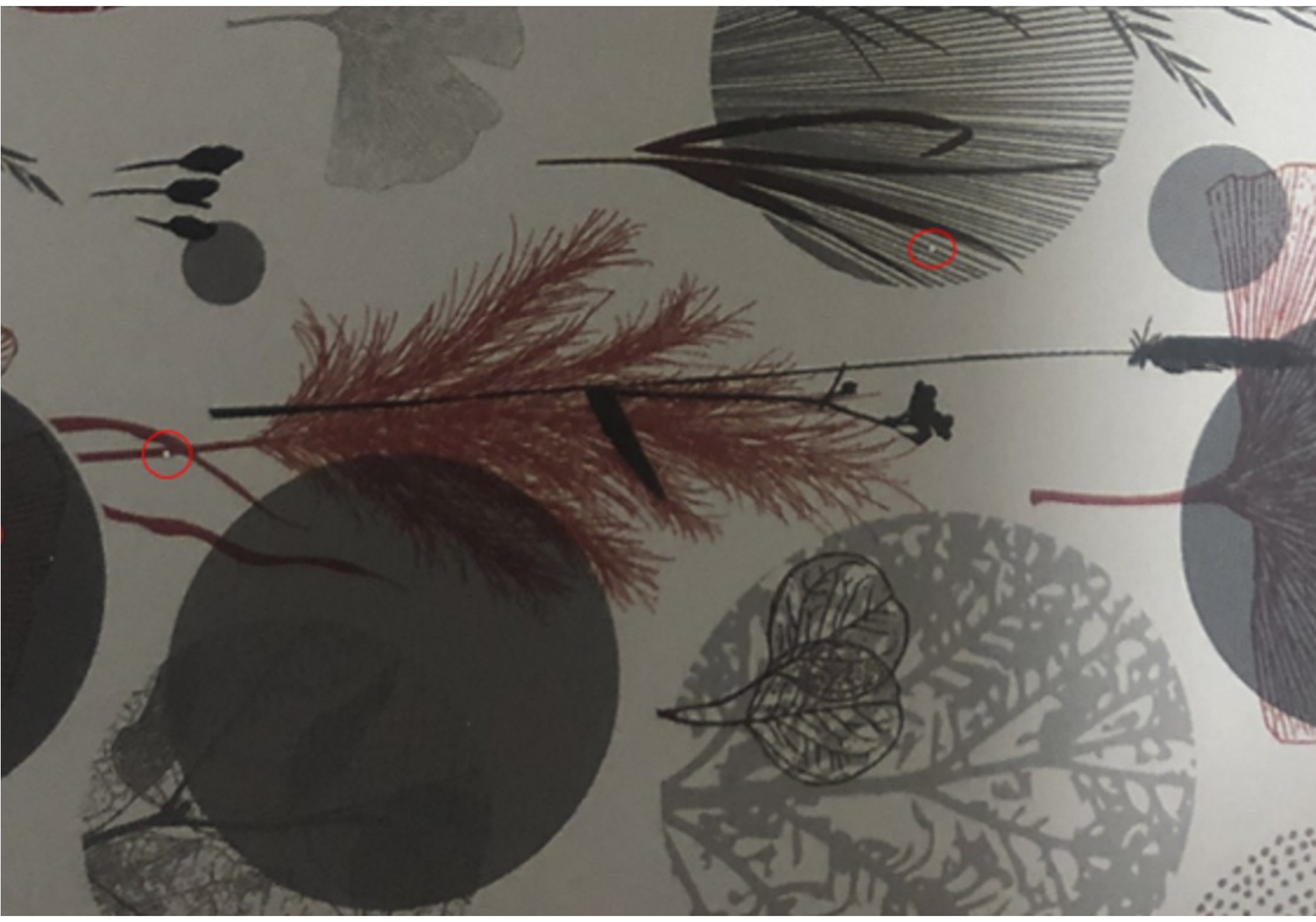}
            \caption{Some fisheye defects bounded by the red circles}
            
            \label{fig:zoom_fisheye}
        \end{figure}

        
        Figure \ref{fig:zoom_fisheye} shows a complex pattern to which white dot defects have been added. What makes this simulation particularly successful is the finesse of the defects (circled in red). They are not coarse white holes, but small imperfections that could easily be missed by an inefficient detection system. The characteristic crater effect can be observed: a very light center (simulating the lack of ink) and a slightly darker edge (the bead of repelled ink). Their random distribution across the surface perfectly mimics contamination by particles such as dust or silicone during the calendering process.
        
\begin{figure}[H]   
            \centering
            \includegraphics[width=0.7\linewidth]{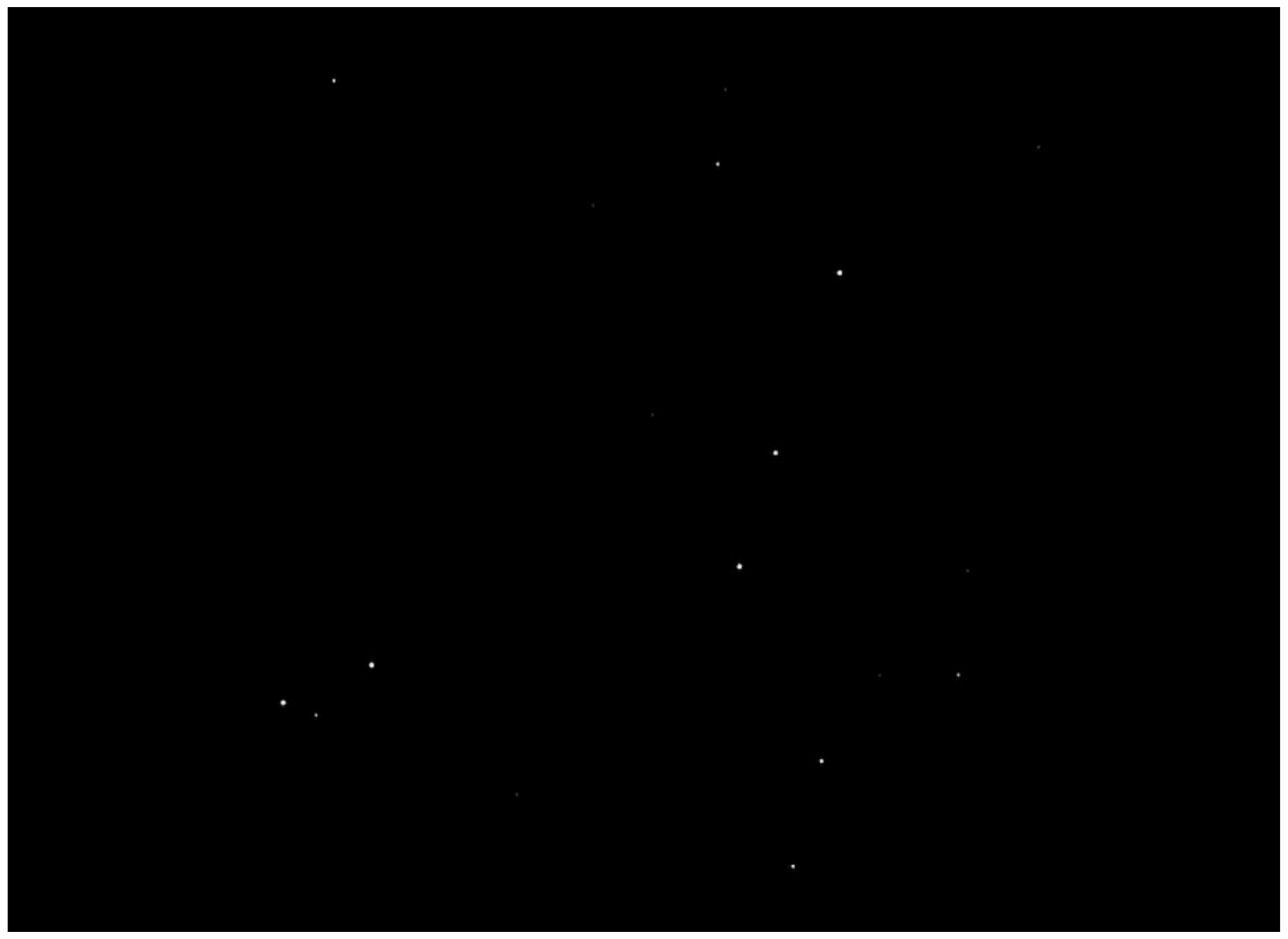}
            \caption{Fisheye mask}
            
            \label{fig:fisheye_m}
        \end{figure}

        Figure \ref{fig:fisheye_m} shows the segmentation mask associated with the defective image in Figure 5. It is a black-and-white binary image that identifies and localises precisely the fisheyes added to the reference image to form the defective image(see Figure \ref{fig:zoom_fisheye}) during the simulation.
        
        \begin{figure}[htbp]
            \centering
            \begin{subfigure}[b]{0.48\textwidth}
                \centering
                \includegraphics[width=\textwidth]{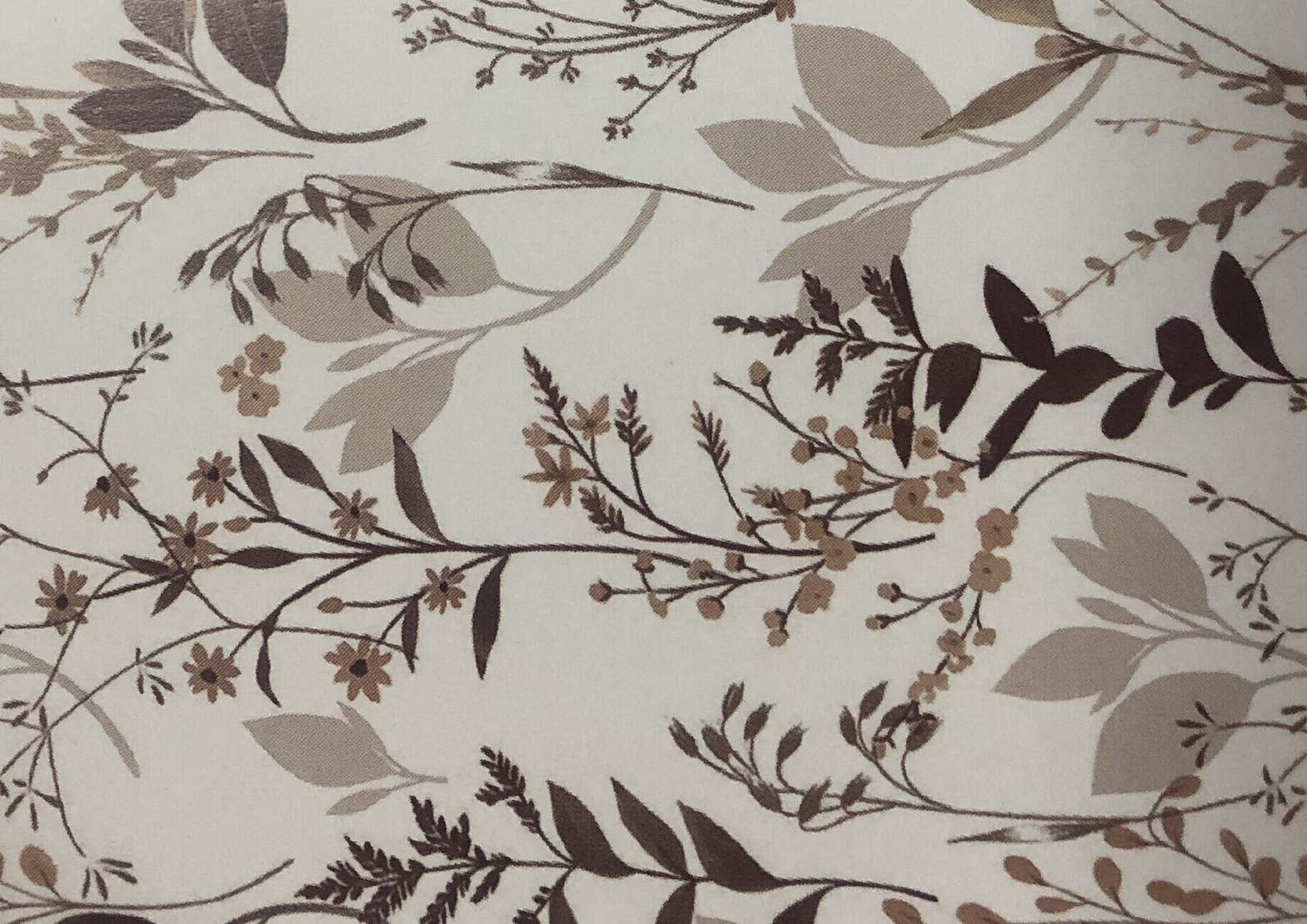}
                \caption{Reference image}
                \label{fig:streak_ok}
            \end{subfigure}
            \hfill 
            \begin{subfigure}[b]{0.495\textwidth}
                \centering
                \includegraphics[width=\textwidth]{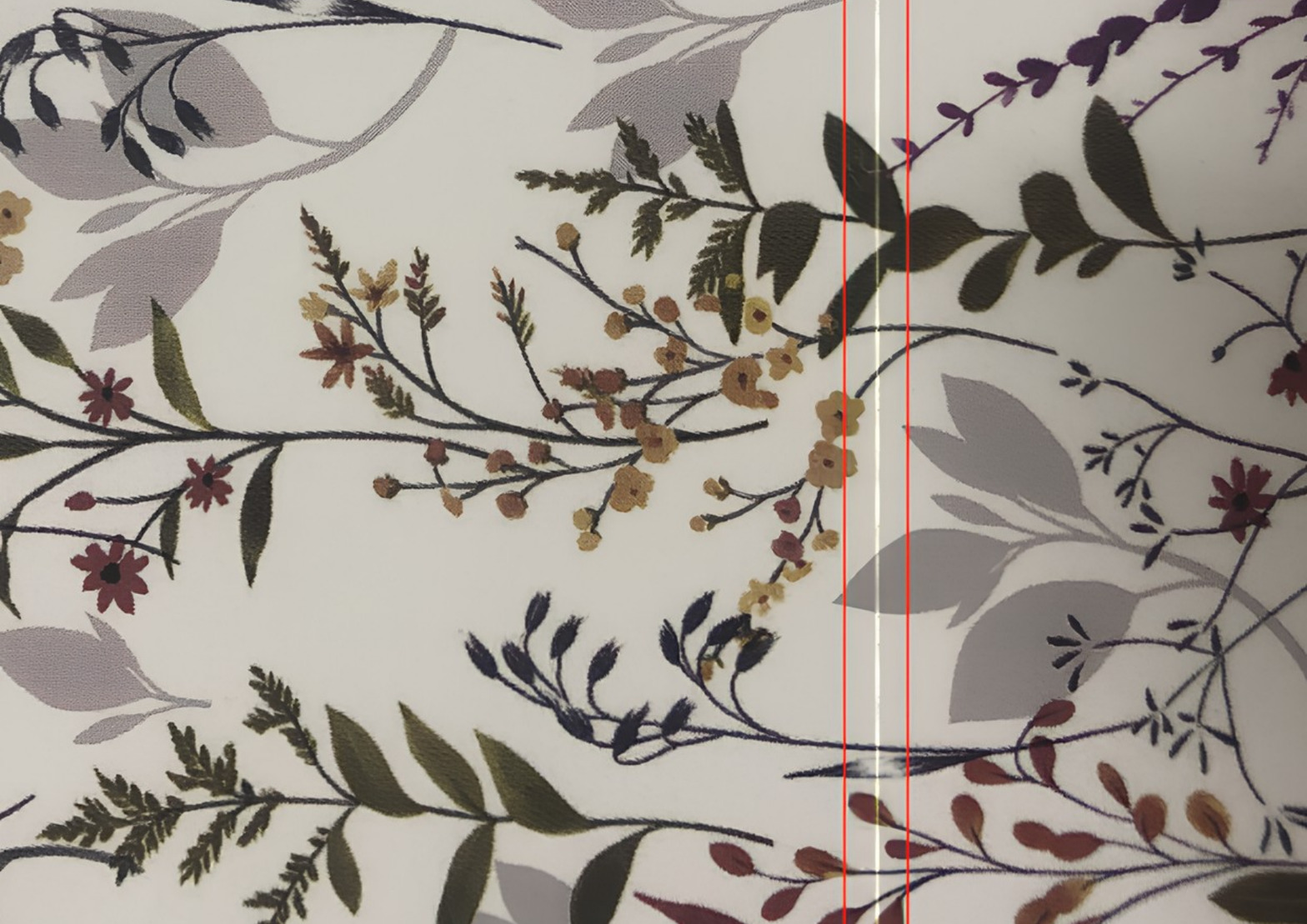}
                \caption{Streak defect, located in the red rectangle.}
                \label{fig:streak_def}
            \end{subfigure}
            
            \caption{Streak defect}
            \label{fig:streak_def_}
        \end{figure}

        The application of algorithm \ref{alg:streak_defect} on Figure \ref{fig:streak_ok} gives the result shown in Figure \ref{fig:streak_def}. This simulation shows a repeating floral pattern on a light background affected by a streak defect, visible as a thin dark or slightly saturated vertical line, bounded by the red rectangle \ref{fig:streak_def}. This narrow, linear, and uniform defect does not alter the pattern itself and indicates an ink deposit anomaly rather than physical deformation. In gravure printing, a streak is an undesirable, continuous or repetitive line or band that is darker (excess ink, smudging, debris) or lighter (lack of ink, obstruction), usually related to the gravure cylinder, squeegee, particles, or ink. The simulation reproduces these industrial characteristics accurately by highlighting the linearity, contrast, uniformity, and vertical orientation of the defect. Our framework also offers the possibility to control these parameters.

        \begin{figure}[H]   
            \centering
            \includegraphics[width=0.6\linewidth]{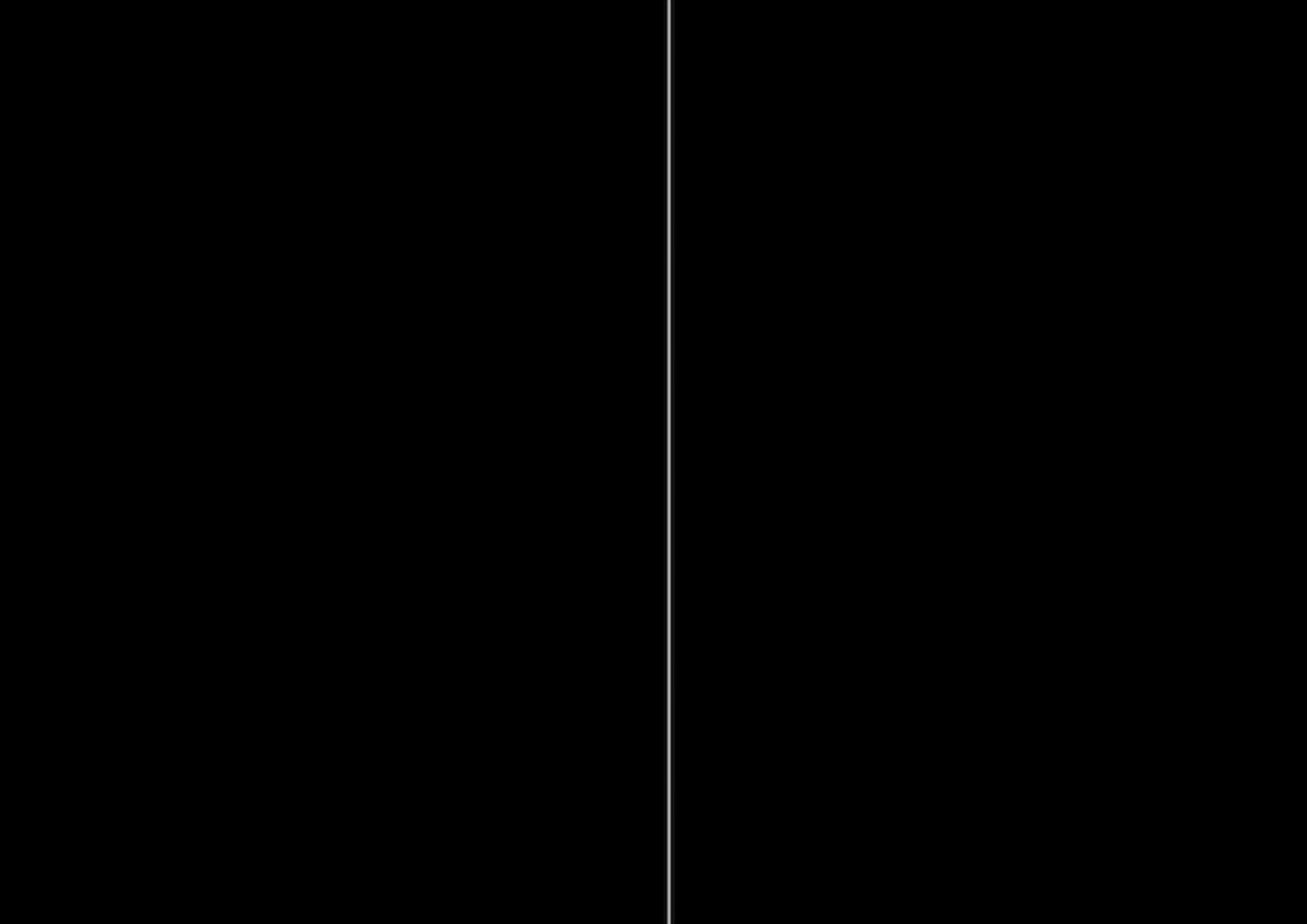}
            \caption{Streak mask, a binary image that isolates the exact area of the streak, confirming its linear nature and precise location.}
            
            \label{fig:streak_m}
        \end{figure}
        
\begin{figure}[htbp]
            \centering
            \begin{subfigure}[b]{0.48\textwidth}
                \centering
                \includegraphics[width=\textwidth]{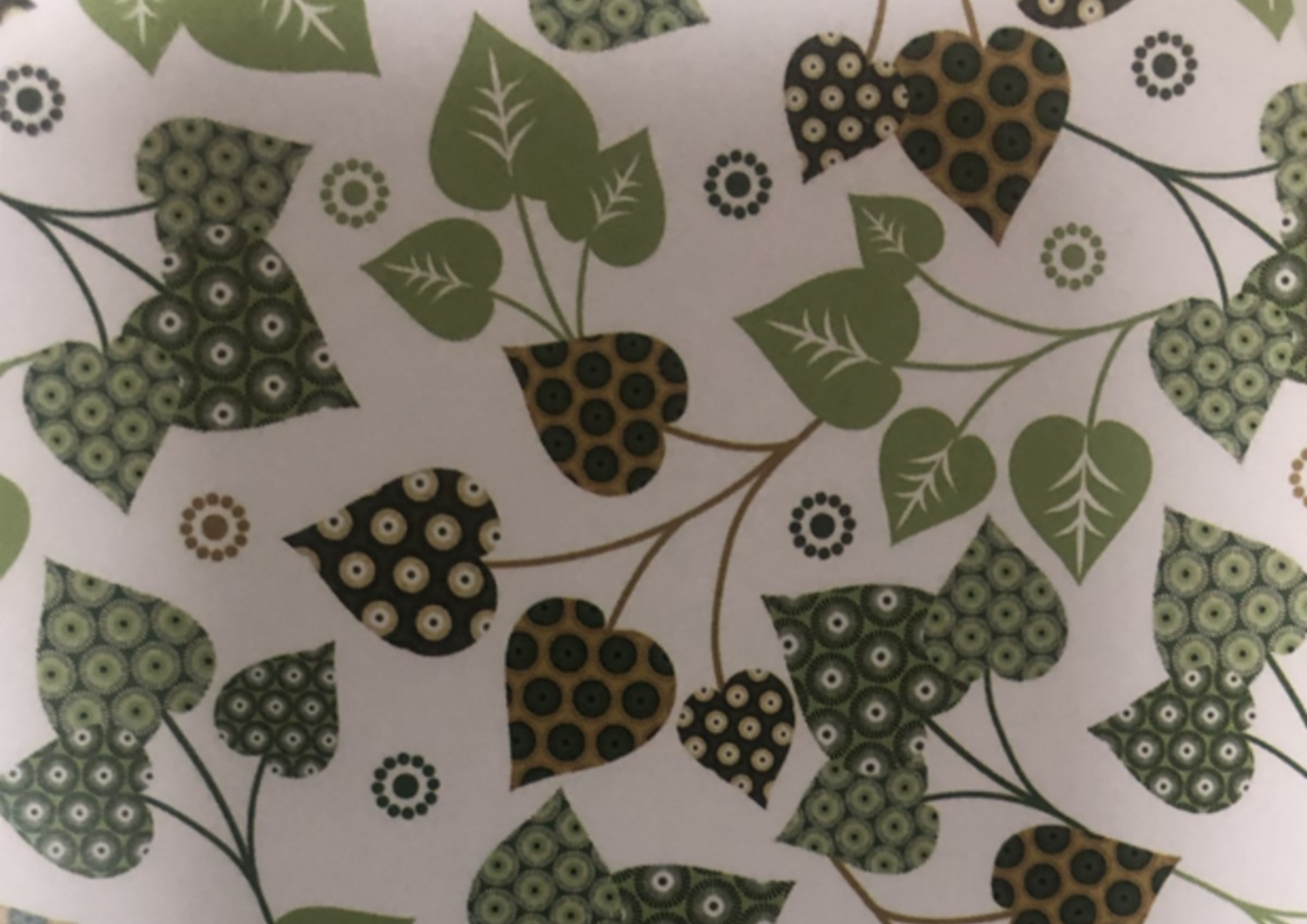}
                \caption{Reference image}
                \label{fig:misregistration_ok}
            \end{subfigure}
            \hfill 
            \begin{subfigure}[b]{0.495\textwidth}
                \centering
                \includegraphics[width=\textwidth]{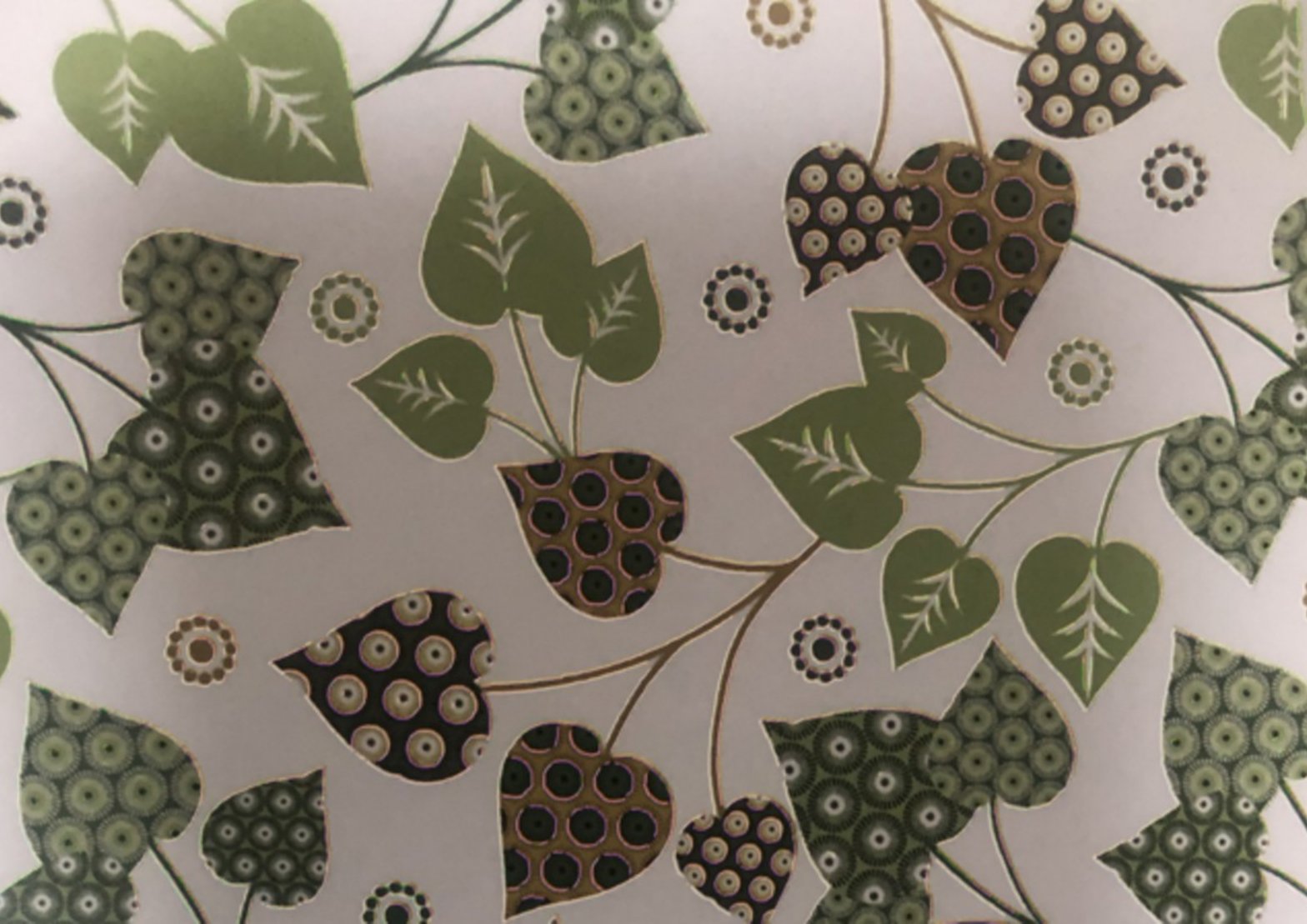}
                \caption{Misregistration defect.}
                \label{fig:misregistration_d}
            \end{subfigure}
            
            \caption{Misregistration defects. Cylinders offset alignment causes incorrect superposition of corresponding colors.}
            \label{fig:misregistration_defect}
        \end{figure}

        The input image in Figure \ref{fig:misregistration_ok} is a clean pattern with no defects. The first step in our method is to extract a precise map of the main contours of the pattern, $M_{\text{contours}}$, using a structured contour detection model. The result of this operation is a binary mask that represents the skeleton of all the shapes present.

        When we apply the misregistration algorithm to Figure \ref{fig:misregistration_ok} we obtain the result shown in Figure \ref{fig:misregistration_d}. For this simulation, the framework selected five cylinders in the order: magenta, black, direct orange, direct green, and varnish/white, with the black cylinder serving as the reference. The algorithm then chose to apply an offset to four of these cylinders, each with its own random translation vector: Direct Green and Magenta were offset by $(-3, 4)$ pixels, Varnish/White by $(3, 5)$ pixels, and Direct Orange by $(-1, 1)$ pixels. Several semi-transparent overlapping color fringes can be observed. A golden fringe, corresponding to the Direct Orange cylinder, is particularly visible on the edges of the patterned sheets. These fringes precisely follow the extracted contours, creating a composite and very realistic ink smear effect, characteristic of poor mechanical alignment of several cylinders on the press.

        \begin{figure}[h]   
            \centering
            \includegraphics[width=0.7\linewidth]{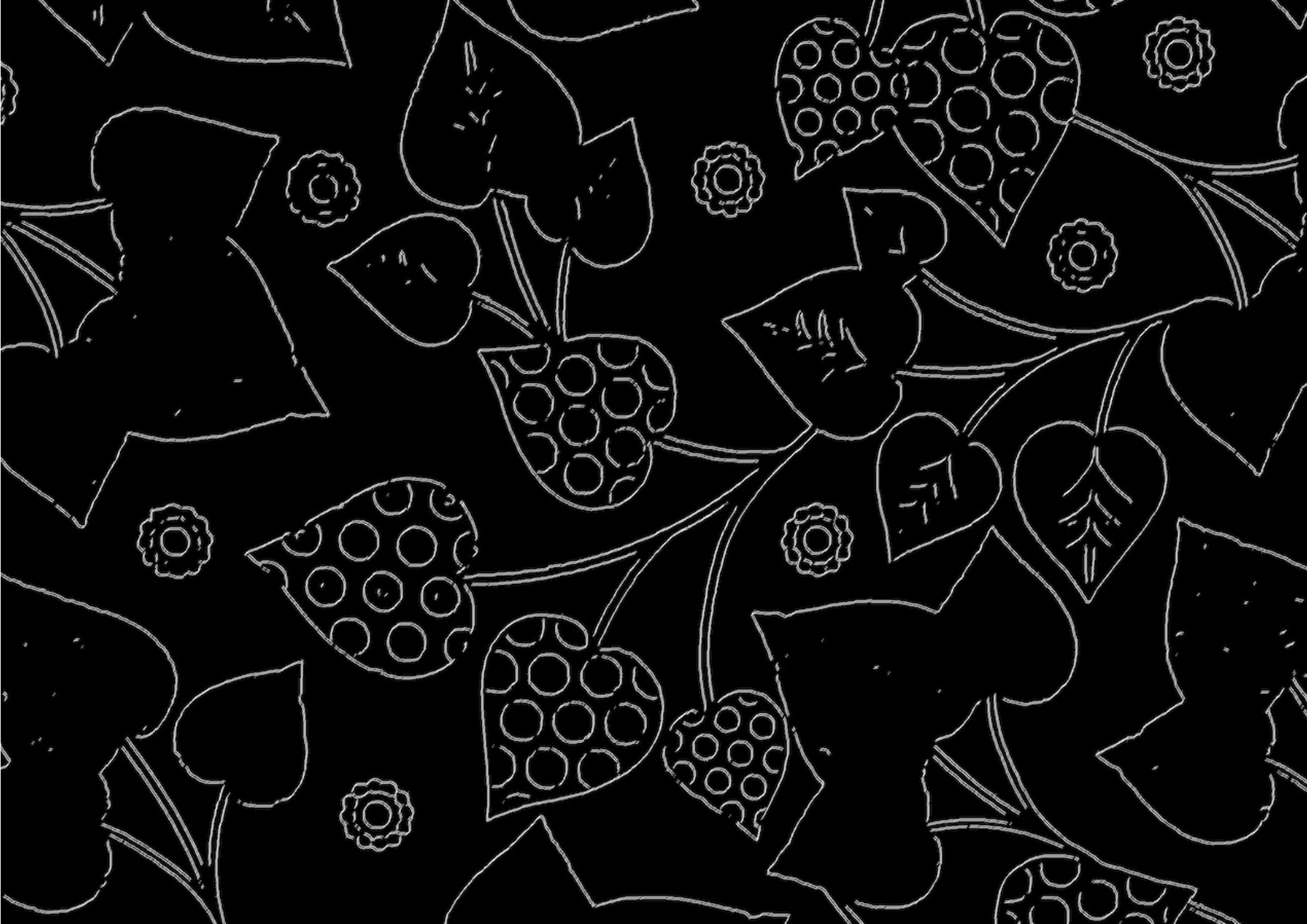}
            \caption{Misregistration defect mask. The mask corresponds to the walls (drawings or outlines) of all the cylinders that were shifted during the simulation.}
            \label{fig:misregistration_mask}
        \end{figure}

        \begin{figure}[htbp]
            \centering
            \begin{subfigure}[b]{0.48\textwidth}
                \centering
                \includegraphics[width=\textwidth]{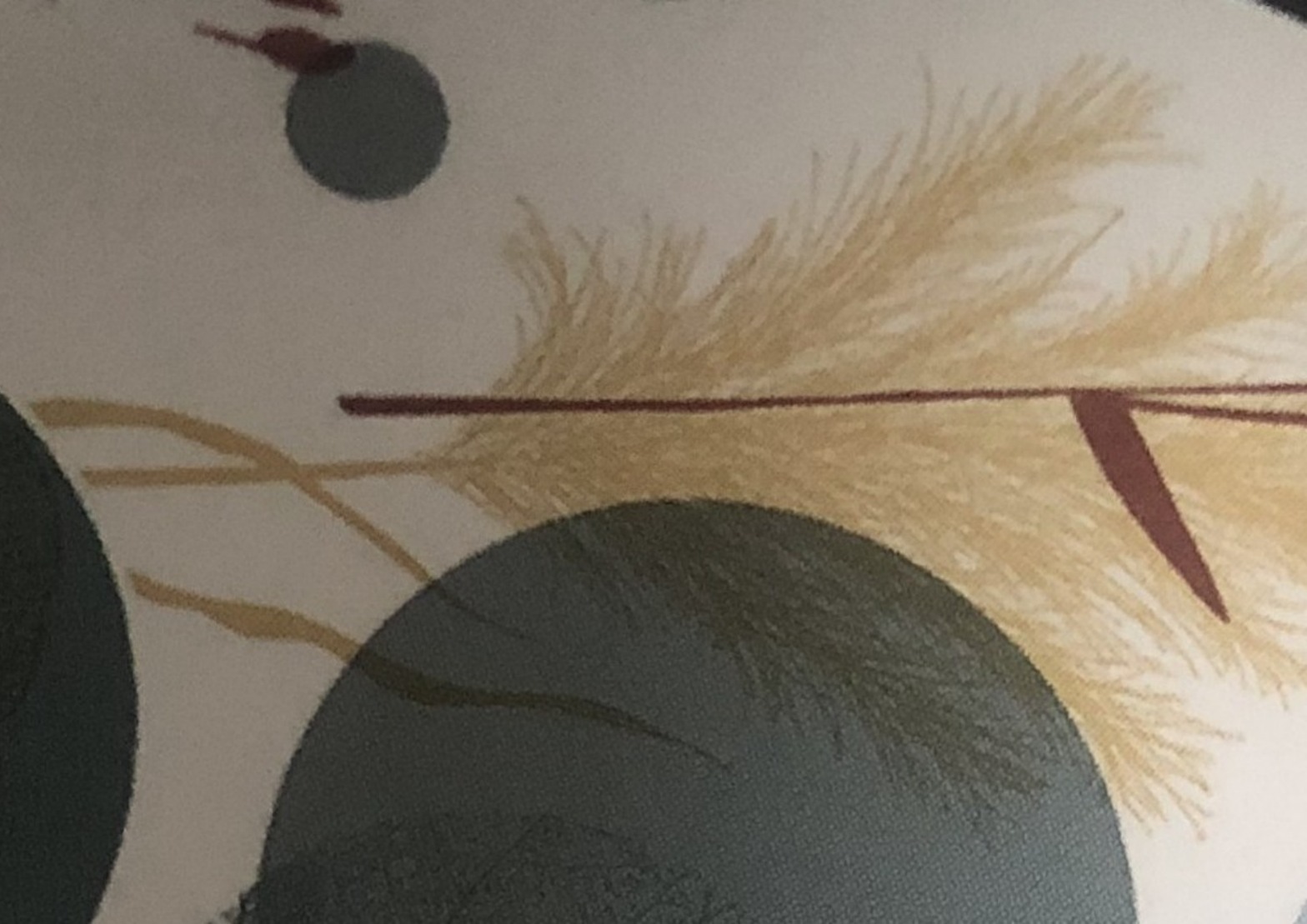}
                \caption{Reference image}
                \label{fig:crease_ok}
            \end{subfigure}
            \hfill 
            \begin{subfigure}[b]{0.495\textwidth}
                \centering
                \includegraphics[width=\textwidth]{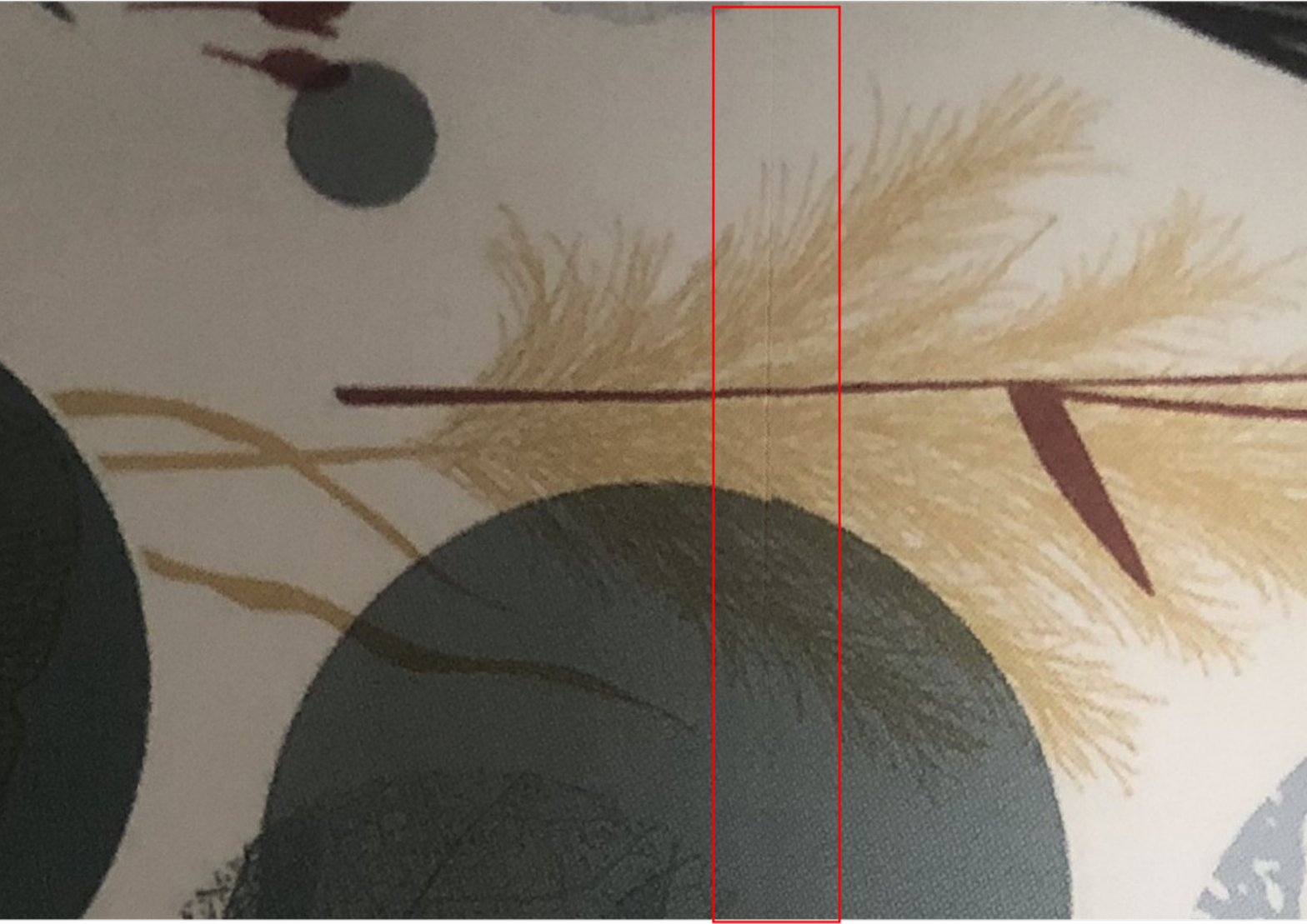}
                \caption{Crease defect, located in the red rectangle.}
                \label{fig:crease_def}
            \end{subfigure}
            
            \caption{Crease defect.}
            \label{fig:crease_defect}
        \end{figure}

        \begin{figure}[H]   
            \centering
            \includegraphics[width=0.7\linewidth]{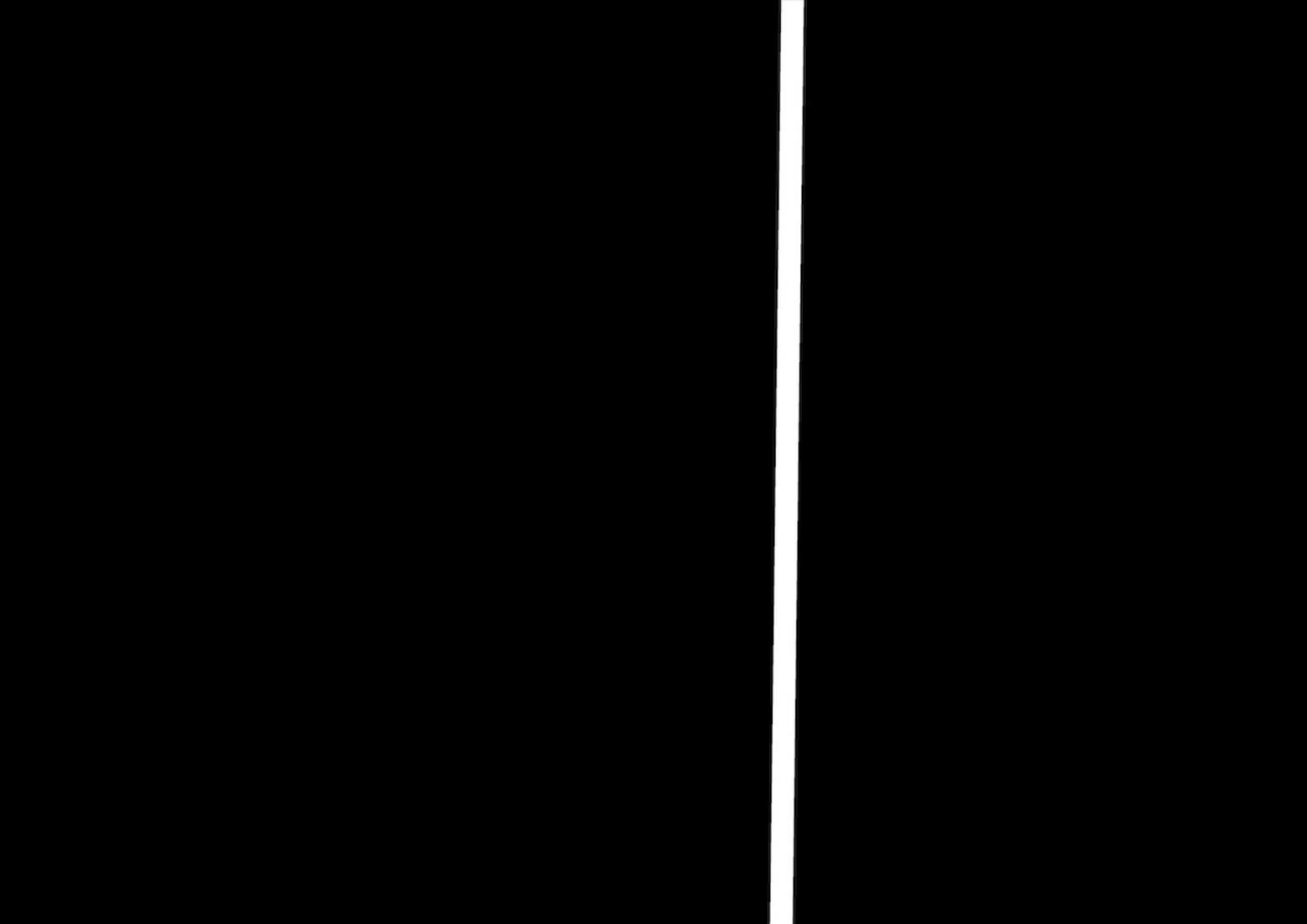}
            \caption{Crease mask defect.}
            \label{fig:crease_def_m}
        \end{figure}

         The corresponding segmentation mask is pertinent. Instead of a fine line, it represents an area of influence with soft edges, capturing the entire extent of the deformation. This (image, mask) pair thus constitutes an ideal field truth, accurately simulating the complexity of a real crease defect.

        \subsection{Quantitative}
    
        To evaluate the effectiveness of the proposed synthetic data generation framework, a comprehensive dataset consisting of 7,533 synthetically generated images was used. The generated dataset was structured and split on the Roboflow platform to ensure a rigorous evaluation process.
        For the defect detection and segmentation task, a state-of-the-art Roboflow RF-DETR Instance Segmentation (Large) architecture\cite{rf-detr} was deployed. The model was trained using the synthetic samples to evaluate how successfully the generated annotations and defect features could transfer to robust industrial vision models. Testing was strictly performed on an independent, real-world test set consisting of authentic defect images captured directly from the production line during live manufacturing operations.

        \begin{table}[h]
        \centering
        \begin{tabular}{l c c c c}
        \hline
        Model Type & Precision & Recall & F1-Score &  mAP@50 \\ 
        \hline
        RF-DETR (Large) & 85.6\% & 78.3\% & 81.7\% & 80.9\% \\ 
        \hline
        \end{tabular}
        \caption{Performance metrics of the RF-DETR Large model trained on synthetic data}\label{tab:metrics}
        \end{table}
    
        The defective image in Figure \ref{fig:crease_def} obtained by applying Algorithm \ref{alg:crease_defect} to the image in Figure \ref{fig:crease_ok}, does not simply show a single line, but presents a convincing geometric distortion (see inside the red rectangle) where the patterns underneath are visibly pinched along a natural, non-perfectly straight trajectory. This distortion is enhanced by subtle photometric effects of light and shadow that give the defect a sense of physical volume.

        The qualitative performance of the framework is further substantiated in Fig.~\ref{fig:inference_evaluation}, which illustrates a representative inference result on a real-world printing line defect. The model successfully detects and segments the boundary anomalies with high confidence, visually confirming the high precision metric of 85.6\% obtained during the testing phase.
        
        \begin{figure}[htbp]
            \centering
            \begin{subfigure}[b]{0.49\textwidth}
                \centering
                \includegraphics[width=\textwidth]{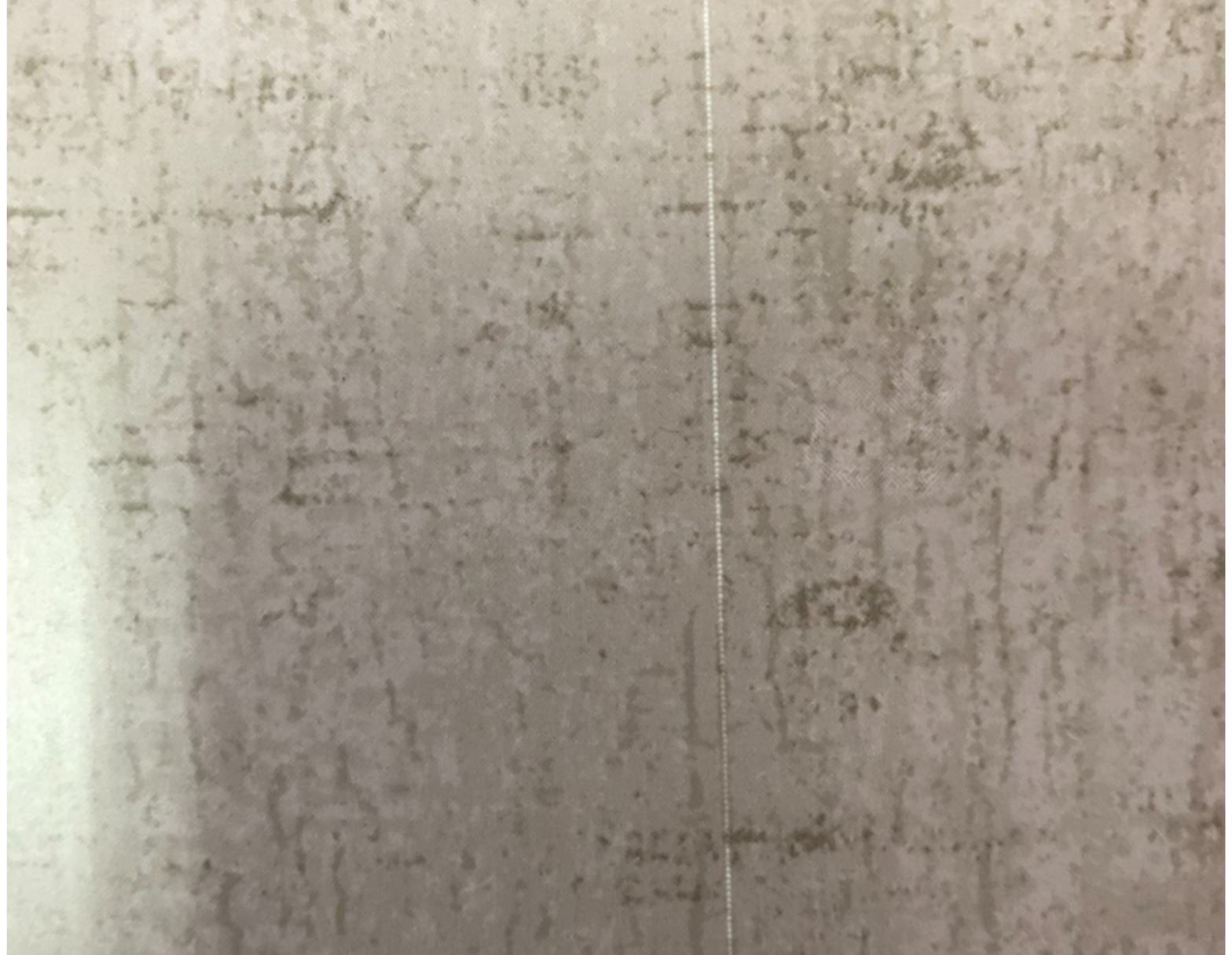}
                \caption{Real production line or streak defect}
                \label{fig:real_defect}
            \end{subfigure}
            \hfill
            \begin{subfigure}[b]{0.49\textwidth}
                \centering
                \includegraphics[width=\textwidth]{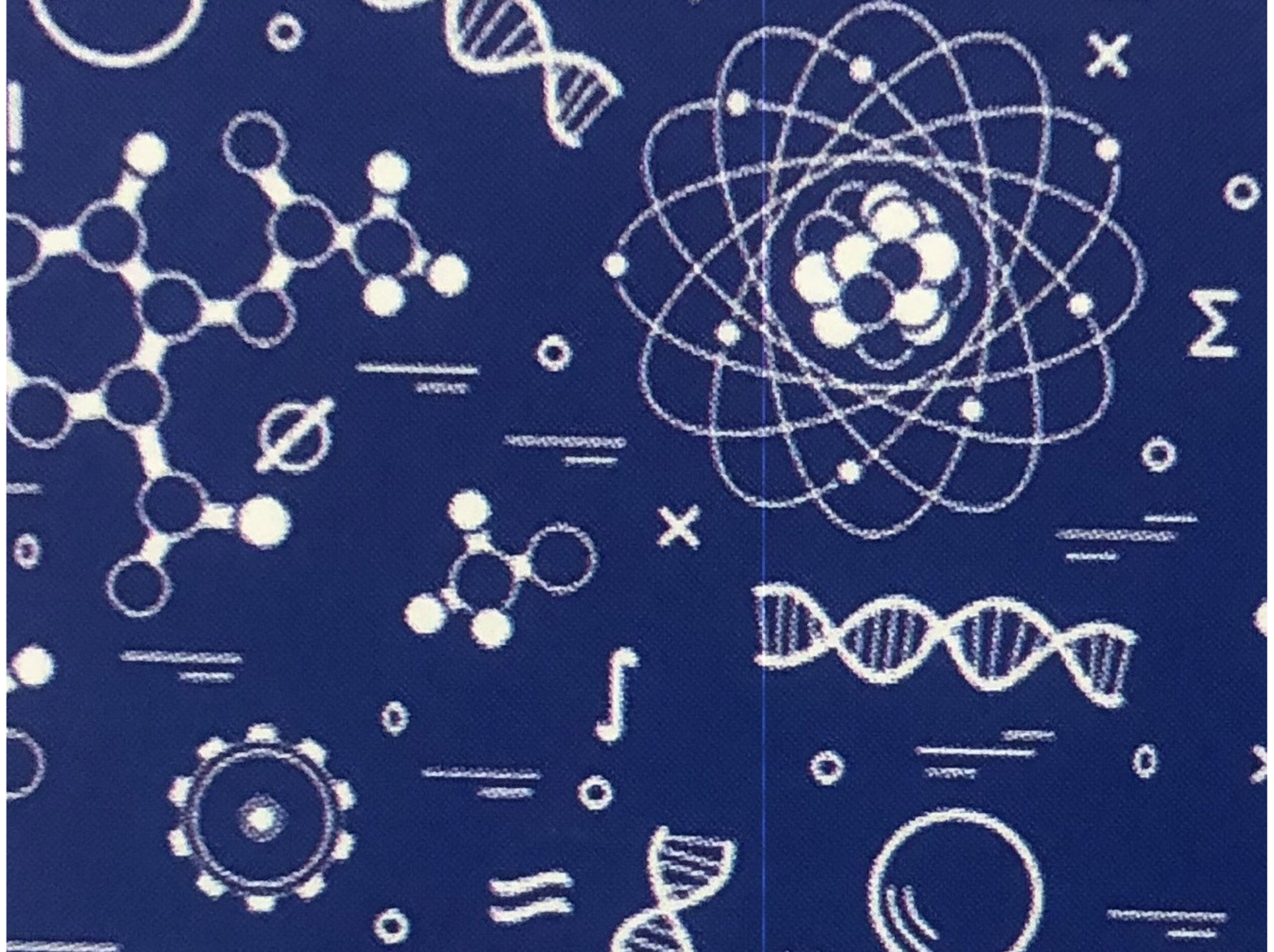}
                \caption{Synthetically generated line or streak defect}
                \label{fig:synthetic_defect}
            \end{subfigure}
            
            \caption{Visual samples from the evaluation dataset: (a) a real  printing streak defect image captured directly from the live production line, and (b) a synthetic streak sample generated by the proposed framework to train the RF-DETR model.}
            \label{fig:dataset_samples}
        \end{figure}
       
        \begin{figure}[H]
            \centering
            \includegraphics[width=1\textwidth]{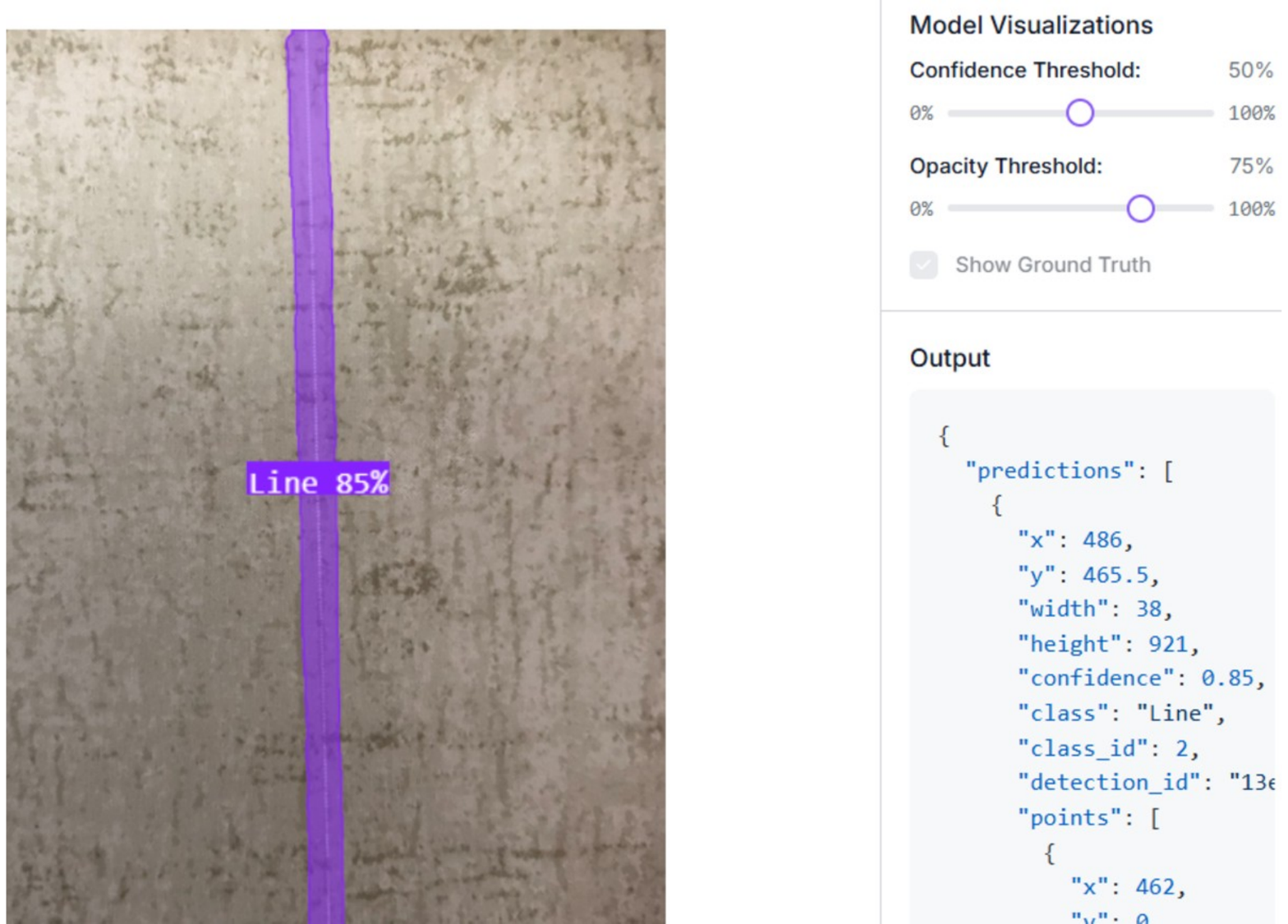}
            \caption{Inference sample of the RF-DETR model evaluated on a real production streak defect. The visualization demonstrates successful instance segmentation and bounding box regression, contributing to the overall test precision of 85.6\%.}
            \label{fig:inference_evaluation}
        \end{figure}

    \section{Discussion}
    
    The results of our synthetic data generation framework, illustrated by simulations of misregistration, crease, streak, and fisheye artifacts, demonstrate the ability of our approach to create a rich, varied, and high-fidelity dataset. Analysis of these simulations shows several key points that validate our methodology.
    Our main objective was not only to create visually realistic defects, but to simulate the physical signature of each anomaly. The crease simulation, for example, does not simply add a line; it applies a nonlinear geometric deformation that pinches the underlying patterns, mimicking the actual stretching of the substrate. Similarly, the simulation of misregistration does not simply move blocks of color, but generates semi-transparent ink fringes on the precise contours of the shapes, which correspond to the effect of imperfect cylinder overlap. This approach, based on modeling physical causes, ensures that the training data exposes the AI model to complex and realistic characteristics, far beyond simple artificial artifacts.
    
    The quality of simulations, in particular for misregistration defects, is based on the use of advanced analysis tools as an intermediate step. By using a structured edge detection model to identify the skeleton of patterns and a CMYK space decomposition to isolate color areas, we ensure that defects are applied consistently with the underlying pattern. A cyan color fringe will only appear on the edges of a shape containing cyan. This intelligence upstream is crucial: it avoids the generation of impossible defects and enhances the credibility of the dataset.
    
    One of the most significant results of our work is the simultaneous generation of perfect segmentation masks. As the figures show, these masks are not simple approximations. The mask of a fold captures its Gaussian influence area, that of a shift delimits the exact shape of the fringe, and that of a fisheye represents the crater. This pixel-level accuracy, achieved without any manual annotation costs, represents a major strategic advantage. Not only does it enable us to bypass the main bottleneck in computer vision projects, but it also provides the model with ground truth of a quality that cannot be achieved by manual means, which is a prerequisite for achieving high performance in segmentation.
    
    By configuring each simulation function, our framework gives us complete control over the diversity of defects generated. We can create subtle, barely visible defects as well as very pronounced ones. We can simulate misalignments on a single cylinder or on several cylinders simultaneously. This ability to generate a variety of scenarios, including extreme cases rarely seen in production, is essential for training a robust model capable of generalizing to unforeseen situations.
    
    The experimental results validate the industrial viability of the proposed framework for automating quality control in rotogravure printing. Achieving a Precision of 85.6\% under strict industry-standard settings proves that the synthetic images generated by our framework provide realistic textures and bounding regions. This high precision minimizes false positives, which is crucial in real-world printing lines to avoid unnecessary machine stops. Furthermore, the model reached an mAP@50 of 80.9\% and an F1-Score of 81.7\%. This robust performance is particularly significant considering that the RF-DETR model was trained entirely on synthetic data and evaluated on a distinct test set. The Recall of 78.3\% indicates that the network successfully learned the diverse geometric variations and color anomalies of printing defects without suffering from severe overfitting. These outcomes confirm that the 7533 synthetically generated samples successfully bridge the reality gap in industrial vision, offering a zero-cost, rapid-deployment solution to bypass the data scarcity bottleneck in smart manufacturing.

    \section{Conclusion}
    
    In this paper, we have presented a comprehensive framework for generating synthetic data, designed to solve the fundamental problem of the scarcity and cost of collecting labeled data for quality control in rotogravure printing. Faced with the difficulty of acquiring examples of real defects and the constant variability of patterns, we have demonstrated that a simulation-based approach is a viable and effective solution that provides a basis for the application of deep learning(computer vision). This work is a robust methodology for modeling the physical signature of a wide range of printing defects. By simulating complex features such as geometric distortion of creases, color fringing from misregistration, splashes, or the texture of streaks, our framework automatically generates defective images and their corresponding segmentation masks(if we want to train a segmentation model) with pixel-level accuracy. The generated datasets can also be adapted to alternative vision models, such as YOLO and RT-DETR.  
    
    The proposed approach enables the low-cost generation of realistic datasets for training computer vision systems. This facilitates the deployment of AI-based quality control solutions in sectors such as packaging, decoration, or printed electronics, where precision, productivity, and defect prevention are essential.

    This study simplifies the modeling of the roll-to-roll printing system for defect modeling. In practice, this type of machine includes a complex set of cylinders, scrapers, motors, and tension mechanisms, which interact to ensure precise and uniform ink deposition \cite{zhaoDynamicSystemModel2022},\cite{seshadriModelingPrintRegistration2013},\cite{kangDesignParameterAnalysis2008a}. Current work focuses on ink deposition, the origin of defects, and the designs found on each cylinder, but does not take into consideration their kinematic parameters, motor speed, or film tension. These results account for the 80.9\% mAP50 accuracy rate achieved by the proposed RT-DETR model on the synthetic dataset. The integration of these factors could be explored in future work.
    
    Consequently, our future work will focus on utilizing this synthetic dataset to train and validate a state-of-the-art segmentation model, specifically YOLOv8-seg. This model will subsequently be deployed in real-world production environments on resource-constrained embedded systems, such as Raspberry Pi and NVIDIA Jetson Orin Nano, for automated quality control of print defects in high-speed rotogravure processes.
    
    \bibliographystyle{elsarticle-num} 
    \bibliography{cas-refs.bib}
    
\end{document}